\documentclass[journal]{IEEEtran}
\usepackage{booktabs}
\usepackage[T1]{fontenc}
\usepackage{graphicx}
\usepackage{amsmath}
\usepackage{amssymb}
\usepackage{array}
\usepackage{url}
\usepackage{cite}
\usepackage{algorithm}
\usepackage{algorithmic}
\usepackage{subfig}
\usepackage{threeparttable}

\begin{document}

\title{\Large Zero-Shot SAM2 Segmentation and Vision Transformer-Based Recognition of Elamite
Cuneiform Symbols from Degraded Tablet Images}

\author{Utsav~Poudel,~\IEEEmembership{}
        Rasik~Bhattarai,~\IEEEmembership{}
        Siddhartha~Pathak,~\IEEEmembership{}
        Raghavendra~Ramacharna,~\IEEEmembership{}
        and~Gaurav~Jaswal~\IEEEmembership{}%
\thanks{U. Poudel is with the School of CS and Engineering, Vellore Institute of Technology, Vellore, and with the Technology Innovation Hub, Indian Institute of Technology, India
(e-mail: utsav.poudel2021@vitalum.ac.in).}%
\thanks{R. Bhattarai is with the College of Technology and Engineering, WestCliff
University, USA (e-mail: rasik.bhattarai@westcliff.edu).}%
\thanks{S. Pathak is with the School of Science \& Technology, York St John University,
London, UK (e-mail: siddhartha.pathak@yorksj.ac.uk).}%
\thanks{R. Ramacharna is with the Department of Information Security and Communication Technology, Norwegian University of Science and Technology, 
Norway, (e-mail: raghavendra.ramachandra@ntnu.no).}%
\thanks{G. Jaswal is with the Technology Innovation Hub, Indian Institute of Technology, India (e-mail: gaurav@tih.iitmandi.ac.in).}}

\maketitle

\begin{abstract}
Automated recognition of ancient cuneiform script poses a compound signal-degradation
problem: the three-dimensional relief of clay tablets creates spatially varying
illumination and cast shadows, surface erosion introduces structured noise that overlaps
with genuine sign impressions, and severe class imbalance across 141 sign categories
undermines classifier reliability. We introduce \textit{EpigraphNet}, a
segmentation-guided transformer pipeline evaluated on the Persepolis Fortification
Archive. From 1,239 annotated tablet images, brightness-adaptive morphological
preprocessing and zero-shot SAM2-Large segmentation generate clean binary symbol masks,
which a fine-tuned Vision Transformer (ViT-B/16) with inverse-frequency class weighting
then classifies. EpigraphNet reaches 86.41\% top-1 accuracy on a 132-class benchmark, a
\textbf{17.21 percentage-point} gain over the strongest CNN baseline (ResNet-101,
69.20\%) and 5.31--12.91\% over four modern backbones (DeiT-B/16, Swin-B, ConvNeXt-B,
EfficientNet-B4) under identical conditions. The full pipeline runs at
$\approx$18\,ms per sign on an NVIDIA A100 GPU. A lower Spearman correlation between
sign frequency and per-class performance indicates more balanced recognition across
frequent and rare classes. Implementation is available at: https://github.com/r11up/sam-guided-vit
\end{abstract}


\section{Introduction}

Recognition of ancient scripts from degraded physical media presents compound
signal-processing challenges distinct from standard document analysis. In clay-tablet
cuneiform, the three-dimensional relief structure of wedge-shaped strokes produces
spatially varying illumination and cast shadows that alter local intensity distributions,
while surface erosion and physical fractures introduce structured noise that overlaps
spectrally with genuine sign impressions. These factors create low signal-to-noise
conditions in which stroke boundaries are poorly defined, a signal-integrity problem
requiring careful preprocessing before any classification stage can succeed. Sign-by-sign
transliteration of Elamite cuneiform requires several days per tablet, and current
practice still relies heavily on manual transliteration and translation, which remain
incomplete and dependent on the knowledge and availability of individual
experts~\cite{bogacz2022}. Automating recognition must overcome both signal degradation
and severe class imbalance across sign categories.

\begin{figure}[t]
    \centering
    \includegraphics[width=\columnwidth]{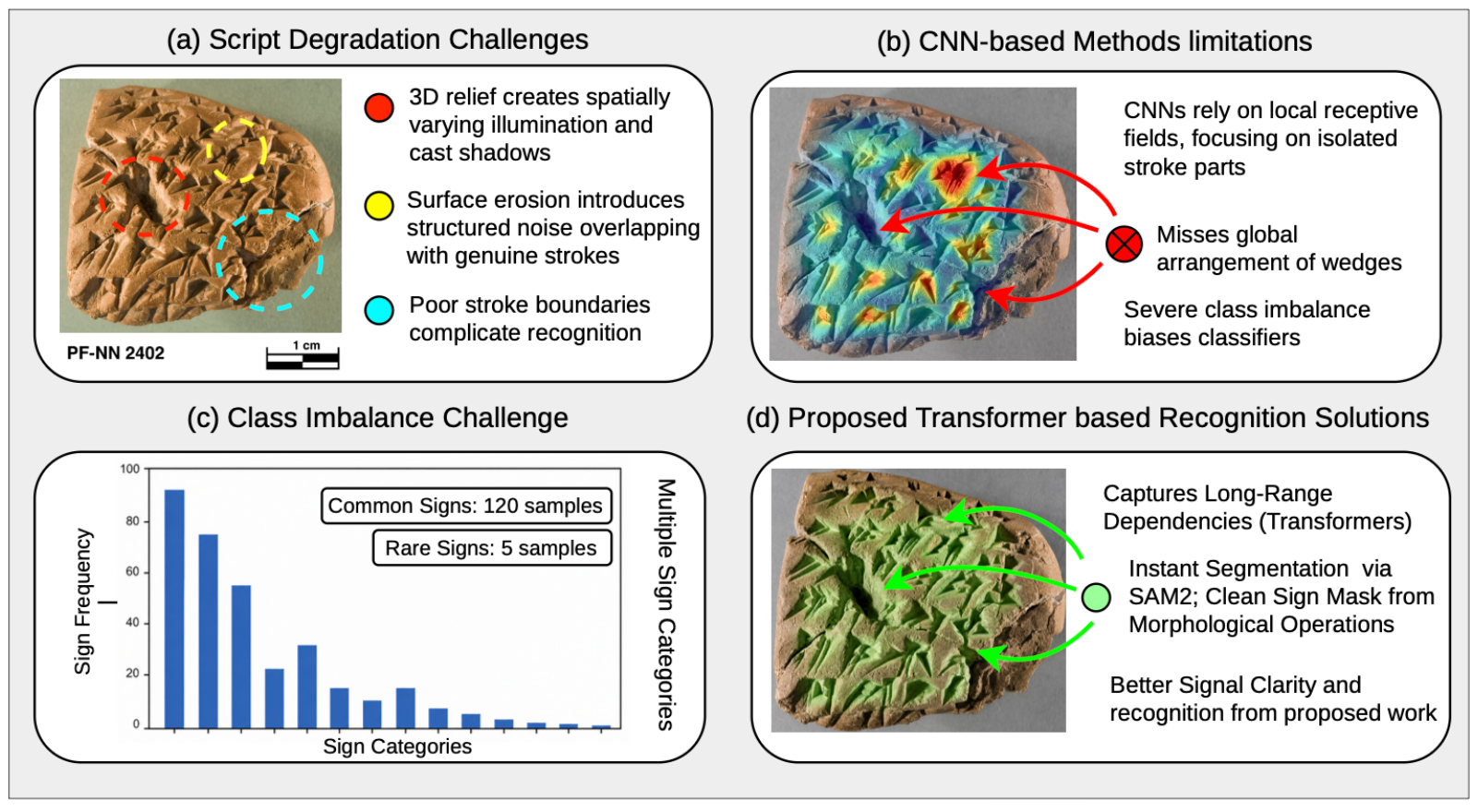}
    \caption{Overview of recognition approaches for degraded cuneiform symbols.
    (a) 3D relief creates spatially varying illumination and
    cast shadows (red)~\cite{hameeuw2024}, surface erosion structured noise
    overlapping (yellow)~\cite{wang2022sts}, and poor stroke
    boundaries (blue)~\cite{khosravy2017} complicate recognition. (b)~CNN-based methods ~\cite{williams2023},missing the global arrangement of wedges that defines each sign~\cite{maath2023}. (c)~Severe class imbalance across 141 sign categories (bar chart)~\cite{lin2017}. (d)~EpigraphNet (ours): SAM2-Large~\cite{ravi2024} with morphological operations and fine-tuned ViT-B/16~\cite{dosovitskiy2020} with inverse-frequency weighting classifies.}
    \label{fig:intro_overview}
\end{figure}

This challenge is not unique to Elamite cuneiform. A broader survey of ancient script
image recognition identifies imbalanced data distribution and image degradation as the
two most persistent obstacles shared across writing systems, from Egyptian hieroglyphs to
Oracle Bone Inscriptions~\cite{diao2025survey}. A wider survey of machine learning for
ancient languages similarly finds that restoration, attribution, and recognition tasks
alike are bottlenecked by scarce, unevenly distributed training examples rather than by
architectural limitations alone~\cite{sommerschield2023}. Synthetic-data pipelines that
simulate erosion, lighting, and surface damage have recently been proposed to counteract
this scarcity for other epigraphic scripts such as Old Aramaic~\cite{aioanei2024},
underscoring that class imbalance and physical degradation must be addressed jointly
rather than in isolation. Fig.~\ref{fig:intro_overview} situates representative deep
learning approaches for degraded ancient-script recognition along these two axes,
motivating EpigraphNet's joint treatment of both problems.

Morphological preprocessing has proven effective for enhancing degraded script features
across OCR tasks under adverse imaging conditions, including handwritten text, number
plates, and printed receipts~\cite{poudel2023}. Translating these filtering principles to
the brightness-variable, three-dimensional relief of cuneiform tablets requires
group-specific parameter tuning, a key signal-processing contribution of this work.

Although deep learning has achieved strong performance in visual recognition, Elamite
cuneiform poses distinct structural challenges~\cite{hameeuw2024}: each sign is defined
by the global arrangement of multiple wedge strokes, favoring architectures that capture
long-range spatial dependencies~\cite{maath2023} over the local receptive fields of
conventional CNNs. Section~\ref{sec:related} reviews cuneiform- and ancient-script-specific
recognition efforts, transformer-based results on other logographic and alphabetic
scripts, and recent foundation segmentation models, situating our contribution relative
to this body of work.

Vision Transformers (ViTs)~\cite{dosovitskiy2020} capture long-range spatial dependencies
via self-attention, making them well-suited for structured symbol recognition. Foundation
segmentation models such as SAM2~\cite{ravi2024} enable zero-shot mask generation from
bounding-box prompts, substantially reducing annotation requirements. We propose
\textit{EpigraphNet}, a segmentation-guided transformer pipeline for automated Elamite
cuneiform sign recognition.

The main contributions of this work are: a SAM2-Large zero-shot segmentation pipeline
using bounding-box prompts that generates clean binary symbol masks without requiring
sign-level mask annotations; and an end-to-end recognition framework integrating
brightness-adaptive morphological enhancement, zero-shot segmentation, and ViT-B/16
classification, with a standardized preprocessing protocol.

\section{Related Work}
\label{sec:related}

\subsection{Cuneiform and Ancient Script Recognition}
Early computational approaches to cuneiform treated sign identification as a retrieval or
detection problem rather than end-to-end classification. Kriege et al.~\cite{kriege2018}
recast sign identification as graph matching over stroke-wedge structure, while Dencker et
al.~\cite{dencker2020} trained a Neo-Assyrian sign detector under weak supervision by
aligning tablet images to existing transliterations, avoiding costly manual bounding-box
annotation. Rest et al.~\cite{rest2022} and St\"{o}tzner et al.~\cite{stotzner2023} instead
addressed the annotation bottleneck through illumination augmentation and 3D-rendering-based
synthetic training data, respectively, and Bucciero et al.~\cite{bucciero2023} extended
3D-rendering supervision to polygonal wedge-level detection. Cobanoglu et al.~\cite{cobanoglu2024}
contributed a large-scale annotated cuneiform sign detection dataset, and Yugay et
al.~\cite{yugay2024} addressed the related but distinct task of stylistic (as opposed to
sign-identity) classification. Closest to our setting, Williams et al.'s
DeepScribe~\cite{williams2023} localizes and classifies Elamite signs from the same
Persepolis Fortification Archive (PFA) used in this work via a RetinaNet-ResNet detector,
and constitutes our primary crop-only CNN baseline (Section~\ref{sec:setup}). Hamplov\'{a}
et al.~\cite{hamplova2024} pursued stroke-level (rather than whole-sign) recognition, and
Simonjetz et al.~\cite{simonjetz2024} and Stelzer~\cite{stelzer2024} addressed downstream
text reconstruction and sign encoding once transliteration is available. Beyond Elamite and
Neo-Assyrian cuneiform, Mahmood et al.~\cite{mahmood2023} classified cuneiform-adjacent
languages using unigram features on a balanced dataset, and Barucci et
al.~\cite{barucci2021} applied a custom CNN (Glyphnet) to Egyptian hieroglyph
classification. Most recently, Elshehaby et al.~\cite{elshehaby2025} trained an ensemble of
five CNN architectures (VGG16, EfficientNet, MobileNet, InceptionResNetV2, and a custom 2D
CNN) on isolated, augmented cuneiform glyphs, reporting near-ceiling accuracy on clean
symbol crops but, unlike the present work, not on unsegmented tablet imagery with natural
class imbalance. Complementing sign-level classification, Mikulinsky et al.'s
ProtoSnap~\cite{mikulinsky2025protosnap} recovers the fine-grained internal wedge-stroke
configuration of a cuneiform sign by aligning a prototype skeleton to the target image via
deep diffusion features, and shows that conditioning synthetic-data generation on these
recovered structures substantially boosts recognition accuracy for rare sign classes in
particular, a data-centric complement to the inverse-frequency weighting and affine
augmentation used in EpigraphNet (Section~\ref{sec:setup}). Bogacz and
Mara~\cite{bogacz2022} survey the wider field of visual cuneiform analysis spanning
manual ink drawings, digital vector graphics, photographs, and 3D scans, and their earlier
geometric neural network approach~\cite{bogacz2020period} addresses period/date
classification of 3D cuneiform tablets, a task complementary to sign-level recognition.
Hameeuw et al.~\cite{hameeuw2024} and Maath et al.~\cite{maath2023} respectively survey
multi-layered tablet visualization for OCR training and classification techniques for
cuneiform imaging broadly, providing complementary context rather than sign-recognition
results directly comparable to Table~\ref{tab:prior_work}.

\subsection{Transformer-Based Recognition of Other Ancient Scripts}
Outside cuneiform, transformer architectures have recently displaced CNN baselines across
several ancient and historical scripts, mirroring the shift motivating our own use of
ViT-B/16. Li et al.~\cite{li2024ancient} combined an improved Swin Transformer with flexible
data-augmentation strategies for ancient Chinese character recognition; Madi et
al.~\cite{madi2024} used multi-task transformer learning for joint Hebrew paleographic script
classification and date estimation; and Surasinghe and Thanikasalam~\cite{surasinghe2026brahmi}
proposed a GAN-transformer framework that generates synthetic Brahmi glyphs to counteract
severe data scarcity before recognition, an imbalance-mitigation strategy conceptually
related to our inverse-frequency weighting and affine-augmentation scheme (Section~\ref{sec:setup}).
A broader survey of deep learning for historical document analysis and recognition by
Lombardi and Marinai~\cite{lombardi2020} corroborates that image degradation and limited or
imbalanced training data recur as central obstacles across historical document types,
including palm-leaf manuscripts, papyri, and other epigraphic media, rather than being
cuneiform-specific quirks. This echoes OCR robustness findings under adverse, non-archival
imaging conditions such as vehicle number plates, receipts, and handwriting~\cite{poudel2023},
lending broader support to the imbalance-aware, noise-robust design choices made in
EpigraphNet.

\subsection{Foundation Segmentation Models for Degraded and Heritage Imagery}
The Segment Anything Model family~\cite{ravi2024} offers prompt-driven, zero-shot mask
generation without task-specific training, making it attractive for heritage domains where
sign-level or object-level masks are expensive to annotate. Outside cultural heritage, SAM2
adaptation typically takes one of two forms: prompting the frozen model directly, as we do,
or parameter-efficient fine-tuning of the encoder for a specialized domain. Xia et
al.~\cite{xia2025mapsam} illustrate the latter with MapSAM, which adapts SAM to historical
map segmentation using low-rank decomposition of the image encoder and automatic prompt
generation, reporting that direct zero-shot SAM application struggles with domain-specific
boundary recognition unless the prompts are of high quality. This finding is consistent with
our own centroid-deviation fallback (Section~\ref{sec:sam2}): rather than fine-tune SAM2's
encoder, we retain a frozen, zero-shot SAM2-Large and instead validate and reject
low-quality masks post hoc, trading some recall on heavily eroded tablets for the ability to
segment without any sign-level mask annotation.

\begin{table*}[t]
\centering
\caption{Comparison of deep learning methods for cuneiform and ancient script recognition.}
\label{tab:prior_work}
\renewcommand{\arraystretch}{1.05}
\footnotesize
\setlength{\tabcolsep}{3pt}
\begin{threeparttable}
\begin{tabular}{
  >{\raggedright\arraybackslash}p{0.20\textwidth}
  >{\centering\arraybackslash}p{0.04\textwidth}
  >{\raggedright\arraybackslash}p{0.16\textwidth}
  >{\raggedright\arraybackslash}p{0.50\textwidth}
}
\toprule
\textbf{Research} & \textbf{Yr.} & \textbf{Task} & \textbf{Method / Key Result} \\
\midrule

\multicolumn{4}{l}{\textit{Cuneiform and Elamite}} \\

Kriege et al.~\cite{kriege2018} 
& 2018 & Sign retrieval 
& Graph-based matching. \\

Dencker et al.~\cite{dencker2020} 
& 2020 & Sign detection 
& Weak supervision using transliteration alignment; mAP evaluation. \\

Rest et al.~\cite{rest2022} 
& 2022 & Sign detection 
& Illumination augmentation; mAP evaluation. \\

Bogacz \& Mara~\cite{bogacz2020period} 
& 2020 & Period classification 
& Geometric neural network on 3D meshes; per-class sample cap. \\

Mahmood et al.~\cite{mahmood2023} 
& 2023 & Language identification 
& DNN: 93.0\%; RF: 95.46\%; balanced dataset. \\

St\"{o}tzner et al.~\cite{stotzner2023} 
& 2023 & Sign detection 
& CNN using 3D renderings; mAP evaluation. \\

Williams et al.~\cite{williams2023} 
& 2025 & Sign classification 
& RetinaNet+ResNet; top-5: 89.0\% (GT crops), 80.0\% (end-to-end). \\

Yugay et al.~\cite{yugay2024} 
& 2024 & Style classification 
& CNN; style accuracy: 83.0\%. \\

Hamplov\'{a} et al.~\cite{hamplova2024} 
& 2024 & Stroke recognition 
& Horizontal stroke detection; accuracy: 90.52\%. \\

Cobanoglu et al.~\cite{cobanoglu2024} 
& 2024 & Sign detection 
& CNN detector; 52K annotated cuneiform signs. \\

Mikulinsky et al.~\cite{mikulinsky2025protosnap} 
& 2025 & Structural alignment 
& Diffusion-feature prototype snapping; improves rare-sign recognition. \\

Elshehaby et al.~\cite{elshehaby2025} 
& 2025 & Sign classification 
& 5-CNN ensemble; EfficientNet accuracy: 99.99\% on clean crops. \\

\midrule
\multicolumn{4}{l}{\textit{3D / Point-Cloud Cuneiform}} \\

Mara \& Bogacz~\cite{mara2019broken} 
& 2019 & Benchmark 
& 3D mesh frontal-alignment normalization; benchmark dataset. \\

Hagelskj\ae{}r~\cite{hagelskjaer2022} 
& 2022 & Point-cloud classification 
& Point-cloud down-scaling network with CNN classifier. \\

\midrule
\multicolumn{4}{l}{\textit{Foundation Models}} \\

Xia et al.~\cite{xia2025mapsam} 
& 2025 & Segmentation 
& SAM + DoRA fine-tuning for historical-map segmentation. \\

\midrule
\multicolumn{4}{l}{\textit{Other Ancient Scripts}} \\

Barucci et al.~\cite{barucci2021} 
& 2021 & Classification 
& Glyphnet CNN for Egyptian hieroglyphs. \\

Demilew \& Sekeroglu~\cite{demilew2019} 
& 2019 & Classification 
& CNN for pre-segmented Ge{\textquotesingle}ez characters. \\

Yue et al.~\cite{yue2022dynamic} 
& 2022 & Classification 
& CNN with dynamic augmentation for Oracle Bone Inscriptions. \\

Wang et al.~\cite{wang2022sts} 
& 2022 & Classification 
& Structure--texture separation network; supports imbalanced data. \\

Li et al.~\cite{li2023longtailed} 
& 2023 & Long-tailed classification 
& CNN + GAN discriminator; Repatch and TailMix augmentation. \\

Li et al.~\cite{li2024ancient} 
& 2024 & Classification 
& Improved Swin Transformer with data augmentation. \\

Madi et al.~\cite{madi2024} 
& 2024 & Classification / dating 
& Multi-task transformer for Hebrew paleography. \\

Zhang et al.~\cite{zhang2024stackedunet} 
& 2024 & Segmentation 
& Stacked-UNets + GAN for multi-script inscriptions. \\

Zhen et al.~\cite{zhen2024yolov8} 
& 2024 & Character detection 
& Improved YOLOv8 with CBAM and small-object head. \\

Hamplov\'{a} et al.~\cite{hamplova2024palmyrene} 
& 2024 & Segmentation / recognition 
& YOLOv8 + CNN with extensive augmentation. \\

Ezhilarasi \& Uma Maheswari~\cite{ezhilarasi2025signarynet} 
& 2025 & Localization / classification 
& SignaryNet CNN with class-balanced augmentation. \\

Surasinghe \& Thanikasalam~\cite{surasinghe2026brahmi} 
& 2026 & Generation / classification 
& BrahmiGAN + PVT/Swin ensemble; 21,195 synthetic glyphs. \\

\midrule
\textbf{EpigraphNet (ours)}
& \textbf{2026} & \textbf{Classification}
& \textbf{SAM2-Large zero-shot segmentation + ViT-B/16; inverse-frequency weighting and affine augmentation; top-1: 86.41\%, top-5: 90.90\%.} \\

\bottomrule
\end{tabular}

\begin{tablenotes}
\footnotesize
\item Note: Comparisons are approximate because datasets, class counts, and evaluation protocols differ. Studies targeting language/style/period classification, structural alignment, 3D point clouds, or non-script segmentation are included for methodological context rather than direct performance comparison.
\end{tablenotes}
\end{threeparttable}
\end{table*}

\subsection{Non-Elamite Cuneiform and Other Ancient-Script Studies with Comparable
Methodology}
\label{sec:non_elamite}
Beyond the Elamite-focused and general-cuneiform literature already discussed, a wider body
of work on \emph{other} cuneiform traditions and \emph{other} ancient scripts shares the
methodological ingredients of EpigraphNet, namely a segmentation or detection stage that
isolates individual glyphs from a noisy, degraded, or three-dimensional surface, followed by
a CNN- or transformer-based classifier, often combined with an explicit strategy for class
imbalance. These studies are collected alongside the Elamite/PFA-focused literature in
Table~\ref{tab:prior_work}, whose segmentation/localization, classifier, and
imbalance-handling details situate EpigraphNet's design choices within this broader body
of work rather than the Elamite/PFA corpus alone.

On the cuneiform side but outside Elamite, Cobanoglu et al.~\cite{cobanoglu2024} released
the largest annotated 2D cuneiform sign-detection dataset to date together with a detection
baseline, directly analogous to the localization stage EpigraphNet inherits from OCHRE
bounding boxes. Mara and Bogacz~\cite{mara2019broken} and Hagelskj{\ae}r~\cite{hagelskjaer2022}
instead operate on 3D tablet scans and point clouds rather than 2D photographs, a
complementary sensing modality to the 2D relief-shadow degradation EpigraphNet addresses,
but one that faces the same annotation-scarcity and class-imbalance obstacles, with Bogacz
and Mara explicitly capping majority classes to counteract imbalance in their period-labelled
subset~\cite{bogacz2020period}.

Outside cuneiform entirely, several studies pair a segmentation or detection front end with
a classifier in a manner structurally close to EpigraphNet's SAM2-then-ViT design.
Hamplov\'{a} et al.~\cite{hamplova2024palmyrene} apply YOLOv8 and Roboflow~3.0 instance
segmentation to isolate individual characters in Palmyrene Aramaic inscriptions before
recognition, mirroring our segmentation-guided extraction of individual signs prior to
classification, albeit for an alphabetic rather than logo-syllabic script. Zhang et
al.~\cite{zhang2024stackedunet} instead perform pure character-mask segmentation,
without a downstream classifier, on self-similar, low-contrast carved stone inscriptions
using a Stacked-UNets/GAN architecture, directly paralleling the mask-extraction role SAM2
plays in EpigraphNet before classification takes over. On Oracle Bone Inscriptions, the
ancient script whose severe long-tailed class distribution most closely parallels the
132-class imbalance in the PFA, Zhen et al.~\cite{zhen2024yolov8} localize characters with
an attention-augmented YOLOv8 detector, Wang et al.~\cite{wang2022sts} separate structural
stroke information from surface-texture noise prior to classification (a structure/texture
decomposition that plays a role similar to our SAM2 mask extraction), Li et
al.~\cite{li2023longtailed} propose adversarial mixup-based augmentation targeted
specifically at tail classes evaluated on the OBC306 and Oracle-20K benchmarks, and Yue et
al.~\cite{yue2022dynamic} propose dynamic dataset augmentation for the same imbalance
problem on rubbing-image crops, all directly comparable in spirit to EpigraphNet's
inverse-frequency loss weighting and affine augmentation of classes with fewer than 50
samples (Section~\ref{sec:setup}). For Ashokan Brahmi, old Tamil, and Grantha
epigraphy, Ezhilarasi and Uma Maheswari~\cite{ezhilarasi2025signarynet} propose a Dynamic
Profiling Bound (DPB) technique for character-level localization on eroded stone
inscriptions followed by a fine-tuned CNN (SignaryNet), explicitly targeting the same
scarce, imbalanced-script setting that motivates Surasinghe and Thanikasalam's synthetic
Brahmi-glyph generation~\cite{surasinghe2026brahmi}. Barucci et al.'s
Glyphnet~\cite{barucci2021} and Demilew and Sekeroglu's Ge'ez
recognizer~\cite{demilew2019} instead apply purpose-built CNNs directly to
already-isolated symbol images drawn from published datasets, in the same
classification-only setting as our crop-only ablation (Section~\ref{sec:ablation}).
Finally, transformer-based recognizers for ancient Chinese characters~\cite{li2024ancient}
and Hebrew paleography~\cite{madi2024} confirm that the ViT/Swin family of architectures
generalizes across logographic and alphabetic ancient scripts alike, reinforcing the
architectural choice underlying EpigraphNet's ViT-B/16 classifier, even though neither
study performs its own image segmentation.

Read across the non-Elamite and other-ancient-script entries of Table~\ref{tab:prior_work},
two patterns emerge that motivate EpigraphNet's
design. First, several classification-only studies on other ancient scripts
(Hebrew paleography~\cite{madi2024}, ancient Chinese~\cite{li2024ancient}, Oracle Bone
adversarial augmentation~\cite{li2023longtailed}, Egyptian hieroglyphs~\cite{barucci2021},
Ge'ez script~\cite{demilew2019}) operate on already-isolated symbol crops from published
benchmarks and report no explicit imbalance-handling mechanism beyond, at most, data
augmentation, leaving open the same segmentation and long-tail problems that motivate our
SAM2-plus-inverse-frequency-weighting design. Second, where a genuine segmentation or
localization stage \emph{is} present, for example Palmyrene instance
segmentation~\cite{hamplova2024palmyrene}, Oracle Bone
detection~\cite{zhen2024yolov8}, stone-inscription character-mask
segmentation~\cite{zhang2024stackedunet}, or Brahmi/Tamil signary
localization~\cite{ezhilarasi2025signarynet}, imbalance is typically handled, if at all, by
data-level augmentation rather than loss-level reweighting integrated with the segmentation
stage itself, as in EpigraphNet. This combination of a segmentation-first pipeline
(Section~\ref{sec:sam2}) with loss-level inverse-frequency weighting
(Section~\ref{sec:setup}) is, to our knowledge, not jointly instantiated in any of the
non-Elamite studies surveyed here, reinforcing the novelty argument made for the
Elamite-specific literature in Table~\ref{tab:prior_work}.

\section{Dataset and Preprocessing}

\subsection{Persepolis Fortification Archive (PFA)}
The Achaemenid Persian Empire (550--330~BCE) was, for over two centuries, the largest
empire the ancient world had yet seen, stretching from the Aegean to the Indus River. Its
day-to-day administration is known almost entirely through a single body of evidence,
the Persepolis Fortification Archive, widely regarded as the most important surviving
primary source for how the empire itself was actually run~\cite{azzoni2017}. The tablets
record the movement of rations, livestock, and travelers across the empire, touching
every social stratum from ordinary laborers to the royal family, and were composed
predominantly in Elamite, a language with no firmly established relation to Old
Persian or any other known language family, yet the empire's principal working script by
virtue of its long-standing scribal tradition~\cite{azzoni2017}. Recovering these records
at scale is therefore not only a pattern-recognition problem but a direct contribution to
reconstructing the logistics and multilingual administration of one of antiquity's
largest states.

The PFA is one of the largest collections of Achaemenid administrative tablets (late
6th--early 5th centuries BCE) in Elamite cuneiform at the Oriental Institute, University
of Chicago. We adopt the publicly available subset in OCHRE~\cite{prosser2023}, which
provides bounding-box labels for sign transliterations. Starting from 1,239 annotated
tablet images spanning 141 distinct sign classes, we discard all classes with fewer than
10 samples, the minimum threshold ensuring at least one representative sample per class
in every stratum of the stratified partition. This removes 9 under-represented classes,
retaining \textbf{132 sign classes} and approximately 18,400 sign crops.

\subsection{Morphological Image Enhancement}
\label{sec:morph}
\begin{figure}[htbp]
    \centering
    \subfloat[Lighter tablet morphological enhancement.]{
        \includegraphics[width=\columnwidth]{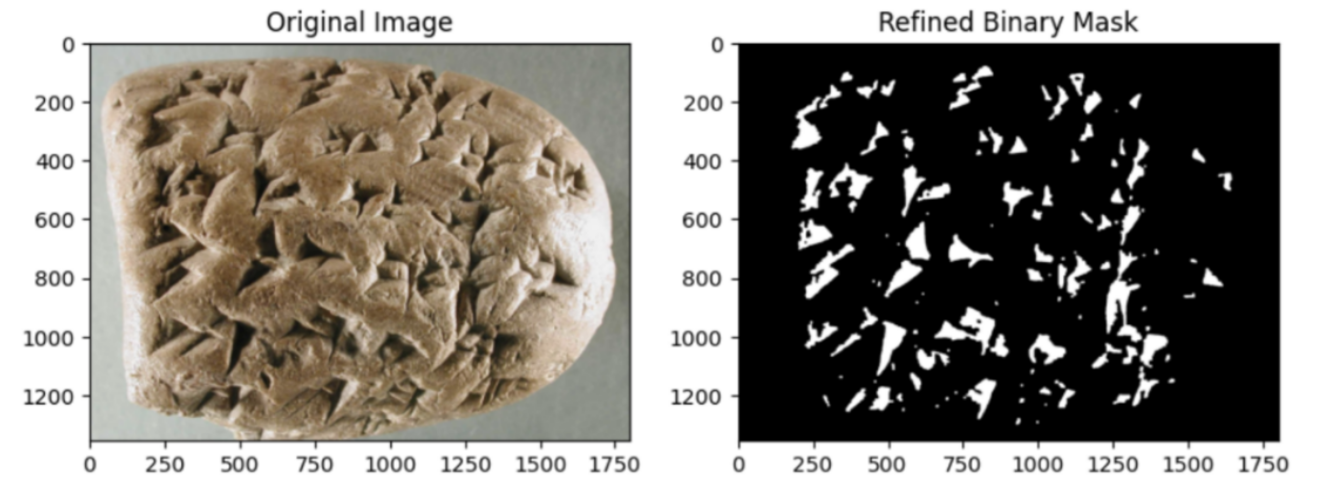}
        \label{fig:morph_light}
    }\\[4pt]
    \subfloat[Darker tablet morphological enhancement.]{
        \includegraphics[width=\columnwidth]{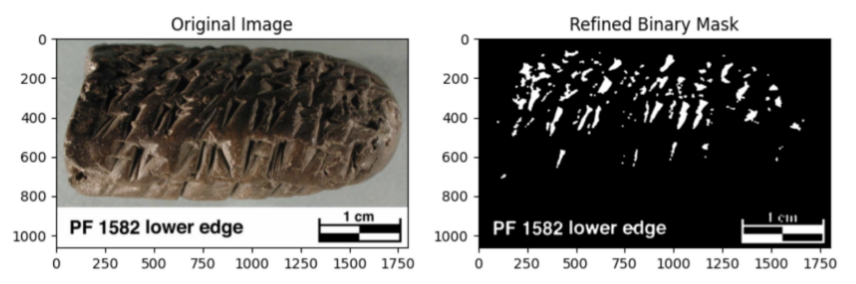}
        \label{fig:morph_dark}
    }
    \caption{Morphological enhancement results for tablets with different brightness levels. (a) Processing pipeline applied to a lighter tablet. (b) Processing pipeline applied to a darker tablet.}
    \label{fig:morph_comparison}
\end{figure}

\begin{figure}[htpb]
    \centering
    \includegraphics[width=\columnwidth]{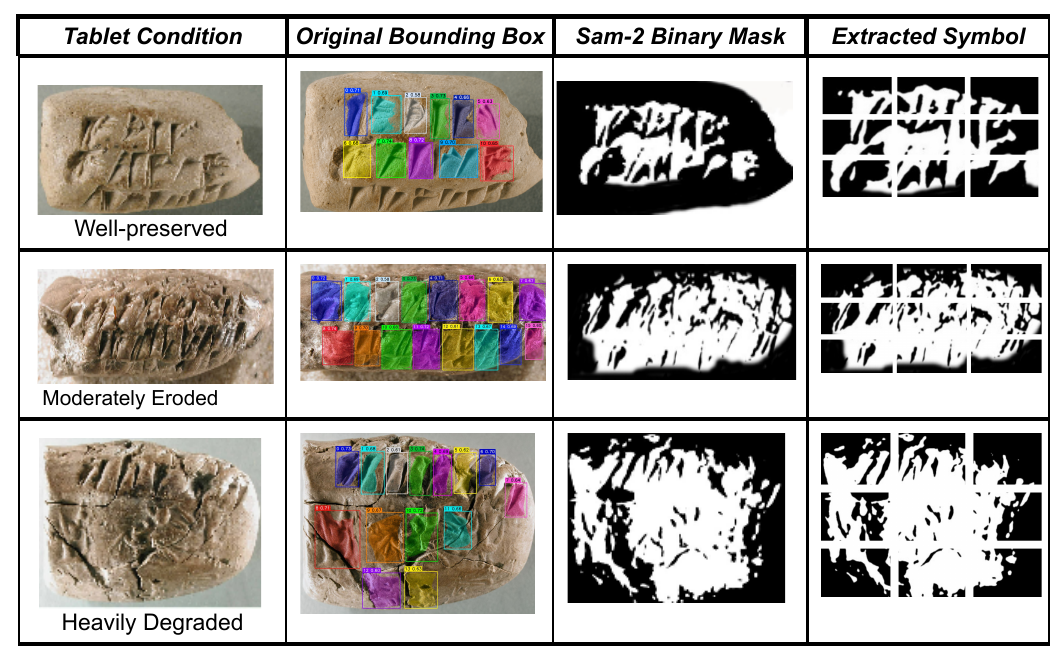}
    \caption{SAM2-Large segmentation across illumination groups (Section~\ref{sec:morph}). Triplets show the bounding box, binary mask, and extracted symbol. Rows 1--2: success cases (well-preserved/eroded). Row 3: fallback to raw crop due to exceeded centroid tolerance $\delta$ on degraded tablets.}
    \label{fig:seg_qualitative}
\end{figure}

Morphological operations enhance sign contrast against the background prior to SAM2
segmentation. For each tablet image, binary thresholding followed by morphological
opening (erosion then dilation) suppresses surface noise, and closing (dilation then
erosion) fills gaps in sign impressions~\cite{khosravy2017}. A uniform threshold is
inapplicable across all tablets owing to significant brightness variation: each image is
assigned to one of three brightness groups (light, moderate, dark) based on mean pixel
intensity and processed with group-specific parameters, and the resulting masks are
merged (qualitative examples appear in Fig.~\ref{fig:seg_qualitative}). Light tablets use
threshold $T_g{=}180$ with a $3{\times}3$ structuring element $B_g$; moderate tablets use
$T_g{=}127$ with $5{\times}5$; dark tablets use $T_g{=}85$ with $7{\times}7$ pixels.
Algorithm~\ref{alg:morph_binary_mask} summarises this group-wise procedure, and
Fig.~\ref{fig:morph_comparison} illustrates representative enhancement results for a
lighter and a darker tablet.

\begin{algorithm}[htpb]
    \renewcommand{\algorithmicrequire}{\textbf{Input:}}
    \renewcommand{\algorithmicensure}{\textbf{Output:}}
    \caption{Morphological Image Enhancement for SAM2 Preprocessing}
    \label{alg:morph_binary_mask}
    \begin{algorithmic}[1]
        \REQUIRE Set of grayscale images $\{I_1, I_2, \ldots, I_n\}$, brightness-group threshold $T$, structuring element $B$
        \ENSURE Enhanced images $\{M_1, M_2, \ldots, M_n\}$ with surface noise suppressed
        \FOR{each image $I_i$ in the dataset}
            \STATE Assign $I_i$ to a brightness group $g \in \{\text{light}, \text{moderate}, \text{dark}\}$ based on mean pixel intensity
            \STATE Select group-specific threshold $T_g$ and structuring element $B_g$
            \STATE Convert $I_i$ to a binary image $I_{bi}$ using threshold $T_g$:
            \[
                I_{bi}(x, y) =
                \begin{cases}
                    1 & \text{if } I_i(x, y) > T_g \\
                    0 & \text{otherwise}
                \end{cases}
            \]
            \textbf{Note:} Thresholding converts the grayscale input $I_i$ to binary image $I_{bi}$; all subsequent morphological operations (Steps 5-6) operate on this binary representation.
            \STATE Apply \textbf{Opening} (Erosion then Dilation) to suppress small noise:
            \[
                I_{\text{opened}} = D(E(I_{bi}, B_g), B_g)
            \]
            \STATE Apply \textbf{Closing} (Dilation then Erosion) to fill small gaps:
            \[
                M_i = E(D(I_{\text{opened}}, B_g), B_g)
            \]
            \STATE Store $M_i$ as the morphologically enhanced image for $I_i$
        \ENDFOR
    \end{algorithmic}
\end{algorithm}

\vspace{-2.5mm}
\subsection{Class Imbalance and Normalization}
The 132-class distribution is severely imbalanced: frequent administrative signs have
hundreds of examples while rare ideographic classes have as few as 10 instances. We
address this via class-specific inverse-frequency weighting of the cross-entropy
loss~\cite{lin2017}. All training images undergo random horizontal flip, random rotation
($\pm 15^{\circ}$), and colour jitter (brightness/contrast $\pm 20\%$). Classes with
fewer than 50 training samples receive additional synthetic copies via random affine
transformations (shear $\pm 10^{\circ}$, scale $0.9$--$1.1$) to reach a minimum of 50
training examples per class. All images are resized to $224{\times}224$ (ViT) or
$1024{\times}1024$ (SAM2) and normalised with ImageNet statistics ($\mu{=}[0.485, 0.456,
0.406]$, $\sigma{=}[0.229, 0.224, 0.225]$). Although derived from natural images, these
statistics maintain compatibility with the ImageNet-21k pretrained ViT-B/16 weights;
since all layers are fine-tuned end-to-end, early-layer representations adapt to the
binary-mask domain during training, mitigating the statistical mismatch. A
\textbf{70/15/15} stratified train/val/test split is used.

\section{Methodology}

\begin{figure*}[htpb]
    \centering
    \includegraphics[width=\textwidth]{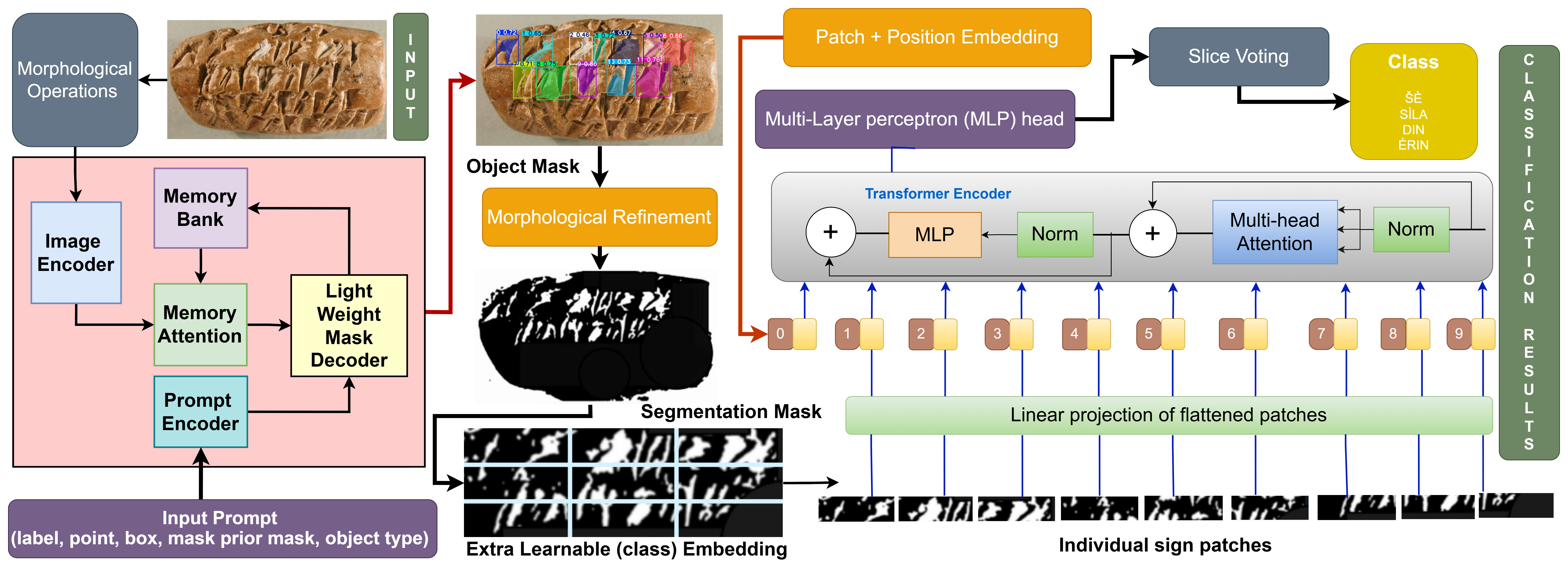}
    \caption{EpigraphNet pipeline: morphological enhancement $\rightarrow$ SAM2-Large
    segmentation $\rightarrow$ mask refinement $\rightarrow$ symbol extraction
    $\rightarrow$ ViT-B/16 classification. \emph{Slice Voting}: the \texttt{[CLS]} token
    aggregates evidence from all 196 patch tokens via multi-head self-attention
    (Section~\ref{sec:vit}).}
    \label{fig:pipeline}
\end{figure*}

\subsection{Pipeline Overview}
EpigraphNet is a five-step pipeline: (1)~brightness-adaptive morphological preprocessing
(Section~\ref{sec:morph}); (2)~SAM2-Large zero-shot segmentation from bounding-box
prompts (Section~\ref{sec:sam2}); (3)~morphological mask filtering (hole-filling,
small-object removal, closing); (4)~background-suppressed symbol extraction; and
(5)~ViT-B/16 classification via patch-level self-attention aggregation
(Section~\ref{sec:vit}). The full pipeline is illustrated in Fig.~\ref{fig:pipeline}.

\vspace{-3mm}
\subsection{SAM2-Large Zero-Shot Segmentation}
\label{sec:sam2}

\begin{figure}[htbp]
    \centering
    \includegraphics[width=0.85\columnwidth]{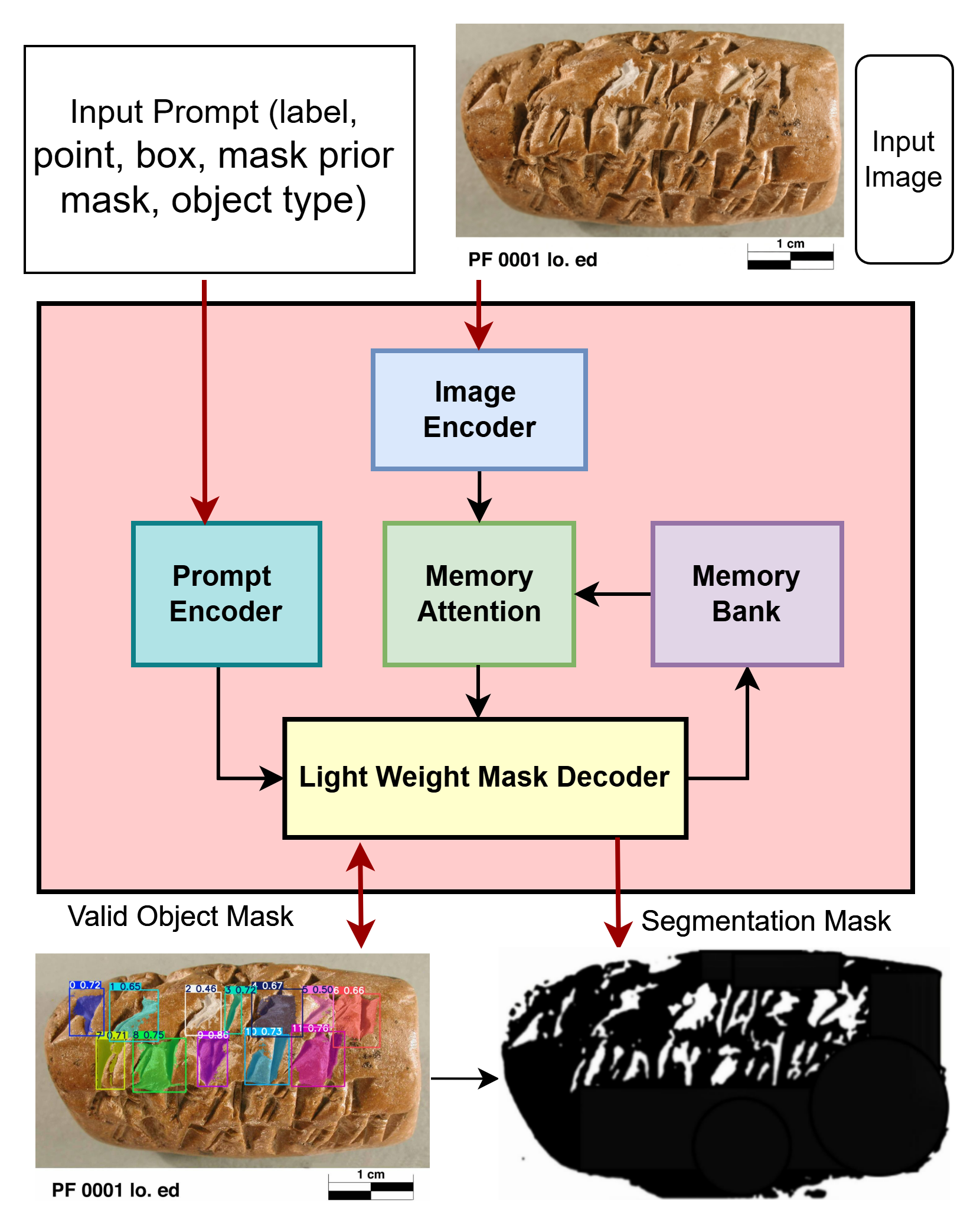}
    \caption{SAM2 model architecture with mask tokens integrated with image features via cross-attention to predict masks of cuneiform symbols.}
    \label{fig:sam2}
\end{figure}

Fig.~\ref{fig:sam2} shows the internal architecture of the frozen SAM2 model used for
segmentation: learned mask tokens attend to image features extracted by the encoder via
cross-attention in the mask decoder, producing candidate masks conditioned on the input
box prompt. For each annotated sign, the OCHRE bounding box $(x_1,y_1,x_2,y_2)$ is passed to the
frozen, pretrained \textbf{SAM2-Large} model~\cite{ravi2024} (312\,M parameters, chosen
over the Tiny/Small/Base+ variants for superior mask quality on low-contrast, degraded
surfaces) as a box prompt; the highest-IoU candidate mask is retained and no SAM2
fine-tuning is performed. Masks are binarised at logit-confidence threshold $\tau{=}0.0$;
holes are filled using \texttt{binary\_fill\_holes}; connected components smaller than
$A_{\min}{=}100$ pixels are removed; and morphological closing suppresses boundary
artefacts. All three hyperparameters ($\tau$, $A_{\min}$, $\delta$) were determined
empirically on a held-out validation subset. Masks whose centroid deviates more than
$\delta{=}20$ pixels from the bounding-box centre are rejected and replaced with the raw
bounding-box crop rather than discarded, ensuring no training or test samples are lost 
a fallback triggered predominantly on heavily eroded tablets where SAM2 latches onto a
neighbouring sign or background texture (Fig.~\ref{fig:seg_qualitative}).
Algorithm~\ref{alg:sam2_seg} formalises this segmentation and refinement procedure.

\begin{algorithm}[htbp]
    \renewcommand{\algorithmicrequire}{\textbf{Input:}}
    \renewcommand{\algorithmicensure}{\textbf{Output:}}
    \caption{Cuneiform Symbol Segmentation via Prompt-Tuned SAM2}
    \label{alg:sam2_seg}
    \begin{algorithmic}[1]
        \REQUIRE Enhanced tablet image $I \in \mathbb{R}^{H \times W \times 3}$, OCHRE bounding box $(x_1, y_1, x_2, y_2)$, threshold $\tau$, area threshold $A_{\text{min}}$, centroid tolerance $\delta$
        \ENSURE Refined binary segmentation mask $M_{\text{final}} \in \{0,1\}^{H \times W}$
        \STATE Resize: $I_{\text{resized}} \leftarrow \text{Resize}(I, (1024, 1024))$
        \STATE Normalize: $I_{\text{norm}} \leftarrow (I_{\text{resized}} - \mu) / \sigma$, where $\mu = [0.485, 0.456, 0.406]$, $\sigma = [0.229, 0.224, 0.225]$
        \STATE Encode bounding box prompt: $Q_{\text{box}} \leftarrow \text{SAM2\_PromptEncoder}(x_1, y_1, x_2, y_2)$
        \STATE Encode image: $F_{\text{img}} \leftarrow \text{SAM2\_ImageEncoder}(I_{\text{norm}})$
        \STATE Decode: $\{(M_k, s_k)\}_{k=1}^{3} \leftarrow \text{SAM2\_Decoder}(F_{\text{img}}, Q_{\text{box}})$
        \STATE Select highest-confidence mask: $M_{\text{raw}} \leftarrow M_{\arg\max_k s_k}$
        \STATE Threshold: $M_{\text{binary}}(x,y) = \mathbb{1}[M_{\text{raw}}(x,y) \geq \tau]$
        \STATE Fill holes: $M_{\text{filled}} \leftarrow \text{binary\_fill\_holes}(M_{\text{binary}})$
        \STATE Remove small objects: $M_{\text{clean}} \leftarrow \text{remove\_small\_objects}(M_{\text{filled}},\, A_{\text{min}})$
        \STATE Suppress boundary artefacts: $M_{\text{final}} \leftarrow \text{morphological\_close}(M_{\text{clean}})$
        \STATE \textbf{Validate:} Compute centroid $C_m$ of $M_{\text{final}}$; reject if $\|C_m - \text{center}(x_1,y_1,x_2,y_2)\|_2 > \delta$
    \end{algorithmic}
\end{algorithm}

\begin{figure*}[t]
    \centering
    \includegraphics[
        width=\textwidth,
        trim={0 0.4cm 0 0.4cm},
        clip
    ]{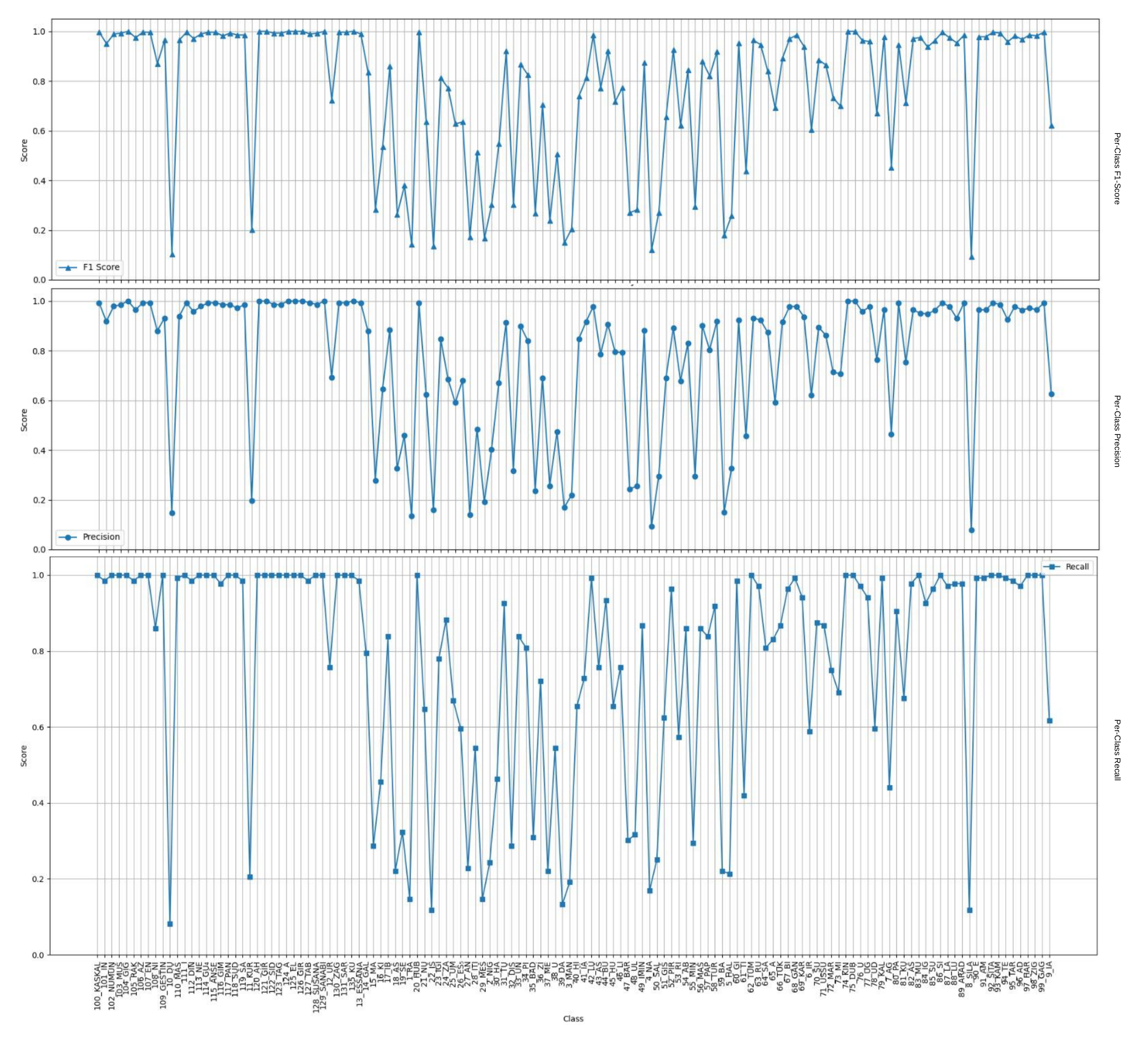}
    \caption{Per-class precision, recall, and F1 of EpigraphNet, sorted by descending class frequency. Class names follow the OCHRE transliteration index; rare signs ($\leq 30$ training samples) appear on the right.}
    \label{fig:perclass}
\end{figure*}

\vspace{-3mm}
\subsection{ViT-B/16 Classification}
\label{sec:vit}
The ViT-B/16~\cite{dosovitskiy2020} partitions a $224{\times}224$ input into a
$14{\times}14$ grid of $16{\times}16$ non-overlapping patches, producing 196 tokens
projected to $d{=}768$ dimensions with learned position embeddings. All 196 tokens
plus a learnable \texttt{[CLS]} token are processed jointly through 12 self-attention
blocks (12 heads each). The \texttt{[CLS]} token aggregates global context from all 196
spatial patch slices, constituting the \emph{Slice Voting} mechanism in
Fig.~\ref{fig:pipeline}: the model weighs evidence from wedge-stroke patterns distributed
across the full symbol rather than merely local regions. The \texttt{[CLS]} representation
is projected to the 132-class output head via a linear classifier. Fine-tuning starts from
ImageNet-21k pretrained weights, with \textbf{all layers fine-tuned end-to-end}
(training hyperparameters are detailed in Section~\ref{sec:setup}). Layer-wise
learning rate decay was evaluated but showed no significant improvement on this
domain-shifted task. Total parameter count: 86\,M.

\subsection{Baseline Architectures}
We train two sets of baselines and four additional modern backbones under identical
experimental conditions (same data split, optimiser, training schedule, and batch size).
\textbf{DeepScribe} (ResNet-18/50/101)~\cite{williams2023,he2016} employs a ResNet backbone
with a softmax classification head on the same 132-class segmentation-prepared crops.
\textbf{YOLOv7}~\cite{wang2023} follows a detection-then-classification approach applied
to full tablet images. Four additional modern architectures —
\textbf{EfficientNet-B4}~\cite{tan2019} (19\,M),
\textbf{ConvNeXt-B}~\cite{liu2022convnext} (89\,M),
\textbf{Swin-B}~\cite{liu2021swin} (88\,M), and
\textbf{DeiT-B/16}~\cite{touvron2021} (86\,M), are evaluated as crop-only classifiers
without the SAM2 segmentation stage, fine-tuned from ImageNet pretrained weights using
the same training protocol as ViT-B/16, providing a comprehensive backbone-level baseline
across the CNN--Transformer spectrum.

\section{Experiments and Results}
\label{sec:results}

\subsection{Experimental Setup}
\label{sec:setup}
Experiments used a single GPU: NVIDIA A100 80\,GiB, PyTorch 2.1 / CUDA 12.1, training with
FP16 mixed precision. SAM2 pre-processing was performed on CPU. Configuration is presented
in Table~\ref{tab:setup}.

\begin{table}[b]
\centering
\caption{Experimental Configuration}
\label{tab:setup}
\renewcommand{\arraystretch}{1.1}
\begin{tabular}{ll}
\toprule
\textbf{Component} & \textbf{Configuration} \\
\midrule
GPU & NVIDIA A100 80\,GB \\
Framework & PyTorch 2.1 / CUDA 12.1 \\
ViT variant & ViT-B/16, ImageNet-21k pretrained \\
Optimizer & AdamW ($\lambda=0.01$) \\
Learning rate & $1\times10^{-4}$, cosine annealing \\
Batch size & 32 \\
Epochs & 50 (5 warmup) \\
Input resolution & $224^2$ (ViT), $1024^2$ (SAM2) \\
Classes & 132 \\
Dataset split & 70/15/15 (train/val/test) \\
\bottomrule
\end{tabular}
\end{table}

ViT-B/16 achieves 12\,ms per-sign inference on the A100; SAM2 preprocessing
($\approx$2.1\,s per tablet on CPU) amortises to $\approx$6\,ms per sign ($\approx$15 signs
per tablet), giving a total pipeline latency of $\approx$18\,ms per sign.

\subsection{Classification Results}

Table~\ref{tab:main_results} reports macro-averaged top-$k$ accuracy on the held-out
132-class test set; macro averaging is used throughout to avoid bias toward the frequent
administrative signs that dominate the class distribution.

\begin{table*}[htpb]
\centering
\caption{Top-$k$ accuracy on PFA test set: crop-only baselines vs.\ full SAM2-Large\,+\,transformer pipeline.}
\label{tab:main_results}
\renewcommand{\arraystretch}{1.1}
\begin{tabular}{lccccc}
\toprule
\textbf{Method} & \textbf{Top-1\,(\%)} & \textbf{Top-3\,(\%)} & \textbf{Top-5\,(\%)}
  & \textbf{Params\,(M)} & \textbf{Time\,(ms)} \\
\midrule
\multicolumn{6}{l}{\textit{Crop-only baselines (no segmentation)}} \\
DeepScribe (ResNet-18)~\cite{williams2023}  & 65.70 & 83.70 & 88.80 & 11.7 & 3 \\
DeepScribe (ResNet-50)~\cite{williams2023}  & 68.30 & 85.10 & 89.70 & 25.6 & 5 \\
DeepScribe (ResNet-101)~\cite{williams2023} & 69.20 & 85.90 & 90.20 & 44.5 & 9 \\
YOLOv7~\cite{wang2023}                     & 76.84 & 83.41 & 87.93 & 36.9 & 12 \\
EfficientNet-B4~\cite{tan2019}              & 73.50 & 83.10 & 87.00 & 19.3 & 5 \\
ConvNeXt-B~\cite{liu2022convnext}           & 76.80 & 84.30 & 87.60 & 88.6 & 9 \\
Swin-B~\cite{liu2021swin}                   & 78.40 & 85.10 & 88.30 & 87.8 & 11 \\
DeiT-B/16~\cite{touvron2021}               & 81.10 & 86.40 & 88.70 & 86.0 & 8 \\
\midrule
\multicolumn{6}{l}{\textit{With SAM2-Large segmentation (full pipeline)}} \\
EfficientNet-B4~\cite{tan2019}    & 79.20 & 85.30 & 88.10 & 331 & 11 \\
ConvNeXt-B~\cite{liu2022convnext} & 81.40 & 86.20 & 88.60 & 401 & 15 \\
Swin-B~\cite{liu2021swin}         & 82.70 & 87.40 & 89.20 & 400 & 17 \\
DeiT-B/16~\cite{touvron2021}      & 84.30 & 88.50 & 89.60 & 398 & 14 \\
\textbf{ViT-B/16 (EpigraphNet)}
  & \textbf{86.41} & \textbf{89.62} & \textbf{90.90} & \textbf{398} & \textbf{18} \\
\bottomrule
\end{tabular}
\end{table*}

\textbf{EpigraphNet vs.\ legacy CNN baselines.}
The 17.21\,pp top-1 improvement over DeepScribe ResNet-101 is consistent with ViT's global
self-attention capturing cuneiform wedge-stroke arrangements more effectively than CNN 
local receptive fields~\cite{dosovitskiy2020}; as the DeepScribe numbers are drawn from
\cite{williams2023} rather than a paired rerun on our split, we treat this comparison as
directional rather than statistically tested. DeepScribe variants achieve competitive 
top-3/5 accuracy but substantially lower top-1, revealing within-family ambiguity 
that global attention resolves. YOLOv7 (76.84\%) outperforms DeepScribe via more precise 
localisation, yet falls 9.57\,pp below EpigraphNet, suggesting that accurate localisation 
alone is insufficient without global feature representation. The value of these background-suppressed 
mask patches is further supported by the ablation study in Section~\ref{sec:ablation}.

Four modern crop-only architectures show a consistent performance hierarchy.
EfficientNet-B4 (73.50\%) provides a parameter-efficient baseline, but its
compound scaling cannot compensate for absent global context. ConvNeXt-B (76.80\%) 
and Swin-B (78.40\%) improve via depthwise convolutions and shifted-window attention, 
yet remain constrained by limited long-range dependency capture. DeiT-B/16 (81.10\%) 
is the strongest crop-only competitor; though architecturally identical to ViT-B/16, 
its CNN-teacher distillation biases it toward local texture, precisely the regime 
insufficient for within-family sign discrimination. The 5.31\,pp gap between DeiT-B/16 
and EpigraphNet reflects this distillation bias and the lack of SAM2 background suppression, 
suggesting that neither architectural modernization nor pretraining alone suffices 
without segmentation-guided signal enhancement. Comparable classification tasks on Oracle 
Bone Inscriptions and other ancient scripts~\cite{fu2022,wang2022sts,fujikawa2023,demilew2019,li2024ancient,madi2024,surasinghe2026brahmi} 
report top-1 accuracies of 84--99.5\%, yet typically use balanced datasets, 
2D manuscripts without 3D-relief noise, and fewer than 100 classes. EpigraphNet's 
86.41\% under a harsher regime of severe class imbalance ($\ge$10 samples), 3D-relief 
degradation, and 132 classes establishes a more demanding, practically relevant benchmark.

\vspace{-3mm}
\subsection{Class-Wise Performance}
\begin{figure}[htpb]
    \centering
    \includegraphics[width=\columnwidth]{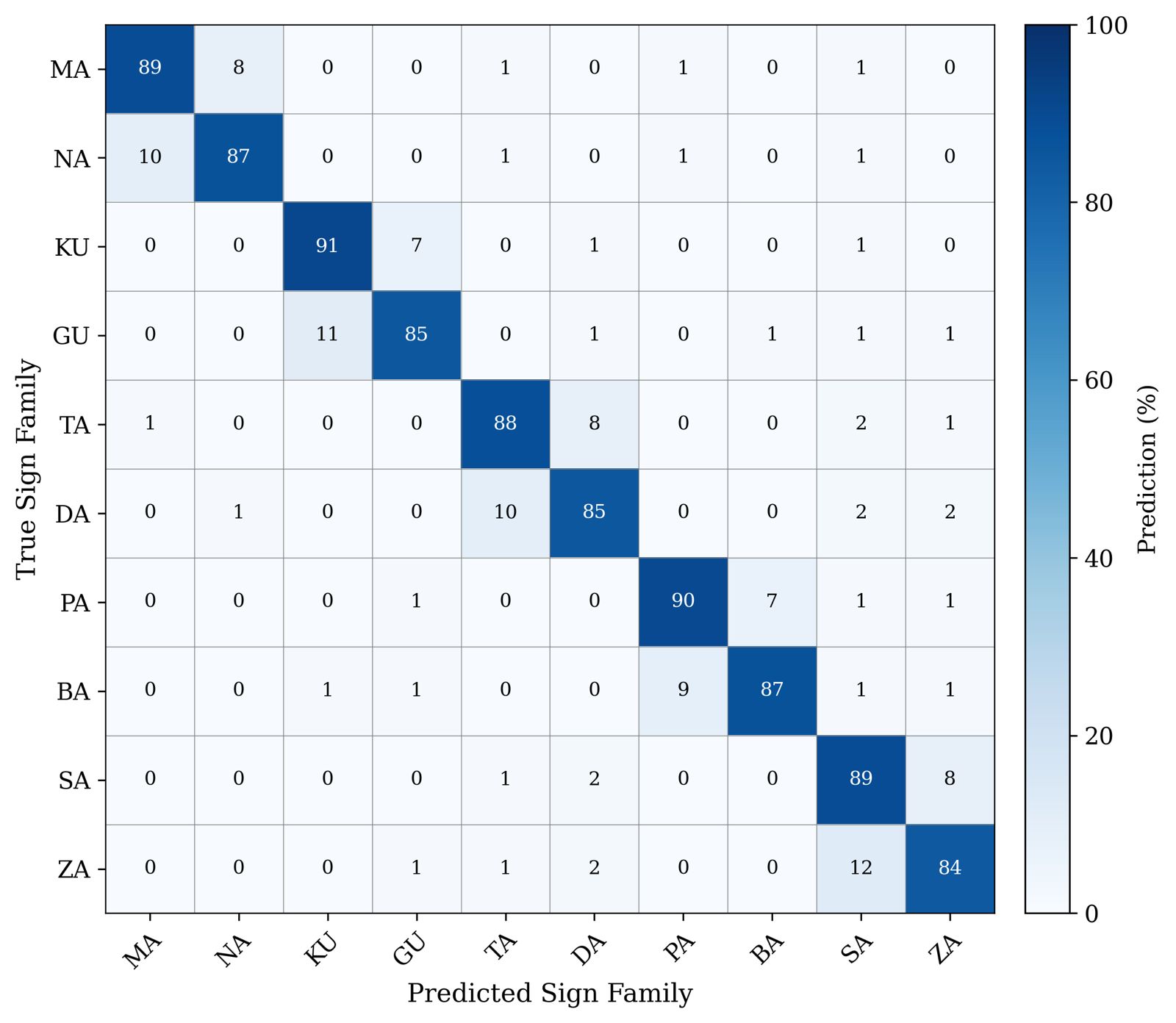}
    \caption{Confusion matrix for the 10 most confused sign pairs. Errors concentrate on
    phonologically and visually similar pairs (e.g., MA/NA, KU/GU)~\cite{elshehaby2022}.}
    \label{fig:confusion}
\end{figure}

The 20 most frequent signs achieve near-perfect precision, recall, and F1, while
accuracy degrades for the least frequent classes owing to limited training data
(Fig.~\ref{fig:perclass}). Misclassifications cluster within phonologically and visually similar sign
families~\cite{elshehaby2022}, as shown in the confusion matrix in
Fig.~\ref{fig:confusion}. Table~\ref{tab:spearman} reports the Spearman~$\rho$
correlation between sign frequency and per-class performance: EpigraphNet achieves
substantially more balanced coverage than the DeepScribe baselines.

\begin{table}[htpb]
\centering
\caption{Spearman $\rho$ Correlation Between Sign Frequency and Per-Class Performance (Lower Indicates More Balanced Recognition Across Frequencies).}
\label{tab:spearman}
\renewcommand{\arraystretch}{1.1}
\begin{tabular}{lccc}
\toprule
\textbf{Architecture} & \textbf{Mean Recall} & \textbf{Prec.\ $\rho$} & \textbf{Rec.\ $\rho$} \\
\midrule
DS RN-18~\cite{williams2023}  & 0.459\,(0.013) & 0.503\,(0.027) & 0.773\,(0.040) \\
DS RN-50~\cite{williams2023}  & 0.483\,(0.018) & 0.509\,(0.049) & 0.794\,(0.042) \\
DS RN-101~\cite{williams2023} & 0.504\,(0.022) & 0.519\,(0.028) & 0.763\,(0.039) \\
\midrule
\textbf{ViT-B/16 (ours)}
  & \textbf{0.623\,(0.016)} & \textbf{0.318\,(0.031)} & \textbf{0.492\,(0.038)} \\
\bottomrule
\end{tabular}
\par\vspace{2pt}

\end{table}

\textbf{Backbone ablation (Table~\ref{tab:main_results}, SAM2 block).}
SAM2-Large preprocessing consistently improves all backbones over their crop-only
counterparts by 4.6 to 5.7\,pp for CNN-based classifiers and 3.2\,pp for ViT-B/16,
indicating that segmentation-guided signal enhancement is complementary to architectural
choices. Within the full pipeline, ViT-B/16 outperforms DeiT-B/16 by 2.11\,pp (86.41\%
vs.\ 84.30\%) despite identical parameter counts (398\,M combined); we attribute this gap
to ImageNet-21k supervised pretraining, which preserves the global attention inductive
bias absent from DeiT's CNN-teacher distillation.

\textbf{Spearman $\rho$ analysis (Table~\ref{tab:spearman}).}
EpigraphNet's $\rho_{\text{prec}}{=}0.318$ and $\rho_{\text{rec}}{=}0.492$ are
substantially lower than DeepScribe variants (0.503--0.519 and 0.763--0.794), indicating
more frequency-independent performance. The $\rho_{\text{rec}}{=}0.492$, a 0.271
reduction from ResNet-101  reflects two cooperative mechanisms: (i) SAM2 background
suppression provides cleaner, more consistent crops for rare classes with only 10--30
training samples; and (ii) inverse-frequency class weighting amplifies gradient updates
for rare classes proportionally, preventing loss domination by the 10--20 most frequent
administrative signs. The lower $\rho$ values do not indicate weaker discrimination of
frequent classes: EpigraphNet's mean recall (0.623) is substantially higher than all
DeepScribe variants (0.459--0.504), indicating improvement across the full class
spectrum. This mirrors the rare-class gains reported by ProtoSnap's structural-prior data
augmentation~\cite{mikulinsky2025protosnap}, suggesting that segmentation-based denoising
and structure-aware synthetic augmentation are complementary, rather than competing,
strategies for closing the frequent/rare performance gap in cuneiform recognition.

\vspace{-5mm}
\subsection{Ablation Study}
\label{sec:ablation}
We isolate the contribution of SAM2-Large segmentation by comparing two configurations
that both retain brightness-adaptive morphological enhancement and inverse-frequency
class weighting, differing only in whether SAM2 segmentation is applied (mean\,$\pm$\,std
over 3 runs). Without SAM2, the ViT-B/16 crop-only classifier reaches
83.20\%\,($\pm$0.41) top-1, 87.00\%\,($\pm$0.36) top-3, and 88.50\%\,($\pm$0.33) top-5
accuracy; adding SAM2 (full EpigraphNet) raises these to 86.41\%\,($\pm$0.38),
89.62\%\,($\pm$0.31), and 90.90\%\,($\pm$0.29), respectively.

Removing SAM2 reduces top-1 by 3.21\,pp but narrows to 2.40\,pp at top-5. Top-5 accuracy
reflects family membership recoverable from stroke texture even with noisy crops, while 
top-1 requires resolving within-family ambiguity where clean SAM2 patches are essential. 
Morphological preprocessing ensures SAM2 receives a high-contrast binary representation; 
without it, dark group tablets ($T_g{=}85$, $B_g{=}7{\times}7$) yield fragmented masks, 
increasing centroid-fallback rates. Ablating class weighting with SAM2 retained reduces 
mean recall from 0.623 to 0.571\,($\pm$0.018), indicating that class weighting contributes
meaningfully to rare-class recall beyond what segmentation alone provides.

\section{Conclusion}

We present EpigraphNet, a segmentation-guided transformer pipeline for automated
recognition of Elamite cuneiform symbols from degraded tablet images.
Brightness-adaptive morphological preprocessing resolves varying tablet illumination 
challenges, while SAM2-Large zero-shot segmentation produces clean binary symbol 
masks without sign-level annotations. A fine-tuned ViT-B/16 achieves 86.41\% top-1, 
89.62\% top-3, and 90.90\% top-5 accuracy on the 132-class PFA benchmark, a 17.21\,pp 
top-1 gain over the strongest CNN baseline (DeepScribe ResNet-101) and a 5.31\,pp gain 
over the best modern transformer alternative (DeiT-B/16). Backbone ablation indicates 
ViT-B/16 is the strongest classifier within our pipeline, and a lower Spearman~$\rho$ 
is consistent with more balanced recognition across frequent and rare classes. At 12\,ms per sign 
classification on an A100 GPU (18\,ms total including amortised SAM2 preprocessing), 
the pipeline is suitable for interactive annotation assistance.

\textbf{Limitations.} EpigraphNet depends on OCHRE bounding-box prompts; unannotated
tablets require an upstream sign-detection stage. SAM2 mask quality degrades on heavily
eroded or fractured surfaces, where the centroid fallback to raw crops applies but cannot
recover fine wedge-stroke structure. CPU-based SAM2 preprocessing ($\approx$2.1\,s per
tablet) limits throughput at scale; a GPU-accelerated preprocessing queue would address
this bottleneck. Reported comparisons against externally sourced baselines (e.g.,
DeepScribe~\cite{williams2023}) are point-estimate accuracy differences rather than
paired statistical tests, since per-sample predictions for those baselines are not
available to us.

\textbf{Future work} will investigate semi-supervised learning on unannotated PFA tablets,
end-to-end integration with large language models for transliteration-to-translation
pipelines, structure-aware synthetic augmentation in the spirit of
ProtoSnap~\cite{mikulinsky2025protosnap} for the rarest sign classes, and extension to
related traditions including Sumero-Akkadian and Neo-Babylonian scripts.

\section*{Acknowledgment}
The authors thank i-Hub and HCI Foundation at IIT Mandi, Himachal Pradesh, India for
supporting this research.

\bibliographystyle{IEEEtran}

\begin{thebibliography}{53}

\bibitem{bogacz2022}
B.~Bogacz and H.~Mara, ``Digital assyriology---advances in visual cuneiform analysis,''
\emph{ACM J. Comput. Cult. Herit.}, vol.~15, no.~2, Art.~38, pp.~1--22, 2022,
doi:~10.1145/3491239.

\bibitem{diao2025survey}
X.~Diao, R.~Bo, Y.~Xiao, L.~Shi, Z.~Zhou, H.~Xu, C.~Li, X.~Tang, M.~Poesio, C.~M.~John,
and D.~Shi, ``Ancient script image recognition and processing: A review,'' \emph{arXiv
preprint arXiv:2506.19208}, 2025.

\bibitem{sommerschield2023}
T.~Sommerschield, Y.~Assael, J.~Pavlopoulos, V.~Stefanak, A.~Senior, C.~Dyer, J.~Bodel,
J.~Prag, I.~Androutsopoulos, and N.~de~Freitas, ``Machine learning for ancient languages:
A survey,'' \emph{Comput. Linguistics}, vol.~49, no.~3, pp.~703--747, 2023,
doi:~10.1162/coli\_a\_00481.

\bibitem{aioanei2024}
A.~C.~Aioanei, R.~R.~Hunziker-Rodewald, K.~M.~Klein, and D.~L.~Michels, ``Deep Aramaic:
Towards a synthetic data paradigm enabling machine learning in epigraphy,'' \emph{PLOS
ONE}, vol.~19, no.~4, p.~e0299297, 2024, doi:~10.1371/journal.pone.0299297.

\bibitem{hameeuw2024}
H.~Hameeuw et al., ``Preparing multi-layered visualizations of Old Babylonian cuneiform
tablets for a machine learning OCR training model,'' \emph{it---Inf. Technol.}, vol.~65,
no.~5, pp.~229--242, 2024, doi:~10.1515/itit-2023-0063.

\bibitem{maath2023}
F.~Maath et al., ``Extensive review of state-of-the-art classification techniques for
cuneiform symbol imaging,'' \emph{Iraqi J. Comput. Sci. Math.}, vol.~4, no.~3,
pp.~116--135, 2023.

\bibitem{barucci2021}
A.~Barucci, C.~Cucci, M.~Franci, M.~Loschiavo, and F.~Argenti,
``A deep learning approach to ancient Egyptian hieroglyphs classification,''
\emph{IEEE Access}, vol.~9, pp.~123438--123447, 2021,
doi:~10.1109/ACCESS.2021.3110082.

\bibitem{mahmood2023}
M.~Mahmood et al., ``Classifying cuneiform symbols using machine learning algorithms
with unigram features on a balanced dataset,'' \emph{J. Intell. Syst.}, vol.~32, no.~1,
p.~20230087, 2023, doi:~10.1515/jisys-2023-0087.

\bibitem{kriege2018}
N.~M.~Kriege et al., ``Recognizing cuneiform signs using graph-based methods,''
in \emph{Proc. Int. Workshop Cost-Sensitive Learning}, PMLR, 2018.

\bibitem{dencker2020}
T.~Dencker et al., ``Deep learning of cuneiform sign detection with weak supervision
using transliteration alignment,'' \emph{PLoS ONE}, vol.~15, no.~12,
p.~e0243039, 2020.

\bibitem{rest2022}
C.~Rest et al., ``Illumination-based augmentation for cuneiform deep neural sign
classification,'' \emph{J. Comput. Cult. Herit.}, vol.~15, no.~3, pp.~1--20, 2022.

\bibitem{williams2023}
E.~C.~Williams et al., ``DeepScribe: Localization and classification of Elamite
cuneiform signs via deep learning,'' \emph{J. Comput. Cult. Herit.}, vol.~18, no.~2,
Art.~31, 2025, doi:~10.1145/3716850.

\bibitem{stotzner2023}
E.~St\"{o}tzner et al., ``CNN based cuneiform sign detection learned from annotated 3D
renderings and mapped photographs with illumination augmentation,''
in \emph{Proc. IEEE/CVF ICCV}, 2023.

\bibitem{bucciero2023}
A.~Bucciero et al., ``R-CNN based polygonal wedge detection learned from annotated 3D
renderings of cuneiform tablets,'' 2023.

\bibitem{yugay2024}
V.~Yugay et al., ``Stylistic classification of cuneiform signs using convolutional
neural networks,'' \emph{it---Inf. Technol.}, vol.~66, no.~1, pp.~17--27, 2024,
doi:~10.1515/itit-2023-0114.

\bibitem{cobanoglu2024}
Y.~Cobanoglu, L.~S\'{a}enz, I.~Khait, and E.~Jim\'{e}nez, ``Sign detection for cuneiform
tablets,'' \emph{it---Inf. Technol.}, vol.~66, no.~1, pp.~28--38, 2024,
doi:~10.1515/itit-2024-0028.

\bibitem{hamplova2024}
A.~Hamplov\'{a} et al., ``Cuneiform stroke recognition and vectorization in 2D images,''
\emph{DHQ: Digital Humanities Quarterly}, vol.~18, no.~1, 2024.

\bibitem{simonjetz2024}
F.~Simonjetz et al., ``Reconstruction of cuneiform literary texts as text matching,''
in \emph{Proc. LREC-COLING 2024}, 2024.

\bibitem{stelzer2024}
D.~M.~Stelzer, ``A recursive encoding for cuneiform signs,''
\emph{it---Information Technology}, vol.~66, no.~6, pp.~232--255, 2024,
doi:~10.1515/itit-2024-0067.

\bibitem{li2024ancient}
J.~Li et al., ``Ancient Chinese character recognition with improved Swin-Transformer and
flexible data enhancement strategies,'' \emph{Sensors}, vol.~24, no.~7, p.~2182, 2024,
doi:~10.3390/s24072182.

\bibitem{madi2024}
B.~Madi et al., ``Multi-task learning for Hebrew paleography: Script classification and
date estimation,'' in \emph{Proc. Int. Conf. Document Analysis and Recognition (ICDAR)},
Springer, 2024, pp.~119--139.

\bibitem{surasinghe2026brahmi}
P.~Surasinghe and K.~Thanikasalam, ``A novel GAN-transformer framework for early Brahmi
script generation and recognition,'' \emph{Eng. Appl. Sci. Res.}, vol.~53, no.~2,
pp.~112--126, 2026, doi:~10.64960/easr.2026.261416.

\bibitem{dosovitskiy2020}
A.~Dosovitskiy et al., ``An image is worth 16$\times$16 words: Transformers for image
recognition at scale,'' in \emph{Proc. ICLR}, 2021.

\bibitem{ravi2024}
N.~Ravi, V.~Gabeur, Y.-T.~Hu, R.~Hu, C.~Ryali, T.~Ma, H.~Khedr, R.~R\"{a}dle,
C.~Rolland, L.~Gustafson, E.~Mintun, J.~Pan, K.~V.~Alwala, N.~Carion, C.-Y.~Wu,
R.~Girshick, P.~Doll\'{a}r, and C.~Feichtenhofer, ``SAM 2: Segment anything in images
and videos,'' in \emph{Proc. Int. Conf. Learning Representations (ICLR)}, 2025.

\bibitem{fu2022}
X.~Fu et al., ``Improvement of oracle bone inscription recognition accuracy: A deep
learning perspective,'' \emph{ISPRS Int. J. Geo-Inf.}, vol.~11, no.~1, Art.~45, 2022,
doi:~10.3390/ijgi11010045.

\bibitem{wang2022sts}
M.~Wang, W.~Deng, and C.-L.~Liu, ``Unsupervised structure-texture separation network
for oracle character recognition,'' \emph{IEEE Trans. Image Process.}, vol.~31,
pp.~3137--3150, 2022, doi:~10.1109/TIP.2022.3165989.

\bibitem{poudel2023}
U.~Poudel et al., ``Applicability of OCR engines for text recognition in vehicle number
plates, receipts and handwriting,'' \emph{J. Circuits Syst. Comput.}, vol.~32, no.~18,
p.~2350321, 2023.

\bibitem{fujikawa2023}
Y.~Fujikawa et al., ``Recognition of oracle bone inscriptions by using two deep learning
models,'' \emph{Int. J. Digit. Humanities}, vol.~5, pp.~65--79, 2023,
doi:~10.1007/s42803-022-00044-9.

\bibitem{demilew2019}
F.~A.~Demilew and B.~Sekeroglu, ``Ancient Geez script recognition using deep learning,''
\emph{SN Appl. Sci.}, vol.~1, Art.~1315, 2019, doi:~10.1007/s42452-019-1340-4.

\bibitem{prosser2023}
M.~Prosser, ``DeepScribe public files,'' Online Cultural and Historical Research
Environment (OCHRE), Nov.~2023. [Online]. Available:
\url{https://pi.lib.uchicago.edu/1001/org/ochre/a3f05985-9cf4-4a39-a4ab-cf51a7ea2d3d}

\bibitem{khosravy2017}
M.~Khosravy, N.~Gupta, N.~Marina, I.~K.~Sethi, and M.~R.~Asharif, ``Morphological
filters: An inspiration from natural geometrical erosion and dilation,'' in
\emph{Nature-Inspired Computing and Optimization}, Modeling and Optimization in Science
and Technologies, vol.~10, Springer, Cham, 2017, pp.~349--379,
doi:~10.1007/978-3-319-50920-4\_14.

\bibitem{lin2017}
T.-Y.~Lin et al., ``Focal loss for dense object detection,''
in \emph{Proc. IEEE ICCV}, 2017, pp.~2980--2988.

\bibitem{he2016}
K.~He, X.~Zhang, S.~Ren, and J.~Sun, ``Deep residual learning for image recognition,''
in \emph{Proc. IEEE CVPR}, 2016, pp.~770--778.

\bibitem{wang2023}
C.-Y.~Wang, A.~Bochkovskiy, and H.-Y.~M.~Liao, ``YOLOv7: Trainable bag-of-freebies
sets new state-of-the-art for real-time object detectors,''
in \emph{Proc. IEEE/CVF CVPR}, 2023, pp.~7464--7475.

\bibitem{tan2019}
M.~Tan and Q.~V.~Le, ``EfficientNet: Rethinking model scaling for convolutional neural
networks,'' in \emph{Proc. ICML}, PMLR, 2019, pp.~6105--6114.

\bibitem{liu2022convnext}
Z.~Liu et al., ``A ConvNet for the 2020s,'' in \emph{Proc. IEEE/CVF CVPR}, 2022,
pp.~11976--11986.

\bibitem{liu2021swin}
Z.~Liu et al., ``Swin transformer: Hierarchical vision transformer using shifted
windows,'' in \emph{Proc. IEEE/CVF ICCV}, 2021, pp.~10012--10022,
doi:~10.1109/ICCV48922.2021.00986.

\bibitem{touvron2021}
H.~Touvron et al., ``Training data-efficient image transformers \& distillation through
attention,'' in \emph{Proc. ICML}, PMLR, 2021, pp.~10347--10357.

\bibitem{elshehaby2022}
M.~Elshehaby et al., ``Cuneiform symbols identification using correlation technique,''
\emph{Iraqi J. Comput. Sci. Math.}, vol.~4, no.~3, pp.~116--135, 2022.

\bibitem{elshehaby2025}
S.~Elshehaby, A.~Panthakkan, H.~Al-Ahmad, and M.~Al-Saad, ``Advanced deep learning
approaches for automated recognition of cuneiform symbols,'' in \emph{Proc. 6th IEEE
Int. Conf. Image Process., Appl. Syst. (IPAS)}, 2025,
doi:~10.1109/IPAS63548.2025.10924496.

\bibitem{lombardi2020}
F.~Lombardi and S.~Marinai, ``Deep learning for historical document analysis and
recognition---A survey,'' \emph{J. Imaging}, vol.~6, no.~10, Art.~110, 2020,
doi:~10.3390/jimaging6100110.

\bibitem{mikulinsky2025protosnap}
R.~Mikulinsky, M.~Alper, S.~Gordin, E.~Jim\'{e}nez, Y.~Cohen, and H.~Averbuch-Elor,
``ProtoSnap: Prototype alignment for cuneiform signs,'' in \emph{Proc. Int. Conf.
Learning Representations (ICLR)}, 2025.

\bibitem{bogacz2020period}
B.~Bogacz and H.~Mara, ``Period classification of 3D cuneiform tablets with geometric
neural networks,'' in \emph{Proc. 17th Int. Conf. Frontiers in Handwriting Recognition
(ICFHR)}, 2020, pp.~246--251.

\bibitem{mara2019broken}
H.~Mara and B.~Bogacz, ``Breaking the code on broken tablets: The learning challenge for
annotated cuneiform script in normalized 2D and 3D datasets,'' in \emph{Proc. Int. Conf.
Document Analysis and Recognition (ICDAR)}, 2019, pp.~148--153,
doi:~10.1109/ICDAR.2019.00032.

\bibitem{hagelskjaer2022}
F.~Hagelskj\ae{}r, ``Deep learning classification of large-scale point clouds: A case
study on cuneiform tablets,'' in \emph{Proc. IEEE Int. Conf. Image Process. (ICIP)}, 2022,
pp.~826--830, doi:~10.1109/ICIP46576.2022.9898032.

\bibitem{hamplova2024palmyrene}
A.~Hamplov\'{a}, A.~Lyavdansky, T.~Nov\'{a}k, O.~Svojše, D.~Franc, and A.~Vesel\'{y},
``Instance segmentation of characters recognized in Palmyrene Aramaic inscriptions,''
\emph{Comput. Model. Eng. Sci.}, vol.~140, no.~3, pp.~2869--2889, 2024,
doi:~10.32604/cmes.2024.050791.

\bibitem{zhen2024yolov8}
Q.~Zhen, L.~Wu, and G.~Liu, ``An oracle bone inscriptions detection algorithm based on
improved YOLOv8,'' \emph{Algorithms}, vol.~17, no.~5, Art.~174, 2024,
doi:~10.3390/a17050174.

\bibitem{zhang2024stackedunet}
P.~Zhang, C.~Li, and Y.~Sun, ``Stone inscription image segmentation based on Stacked-UNets
and GANs,'' \emph{Discov. Appl. Sci.}, vol.~6, Art.~550, 2024,
doi:~10.1007/s42452-024-06264-8.

\bibitem{ezhilarasi2025signarynet}
S.~Ezhilarasi and P.~Uma Maheswari, ``Leveraging digital acquisition and DPB based
SignaryNet for localization and recognition of heritage inscription palaeography,''
\emph{npj Herit. Sci.}, vol.~13, Art.~362, 2025, doi:~10.1038/s40494-025-01913-6.

\bibitem{li2023longtailed}
J.~Li, Q.-F.~Wang, K.~Huang, X.~Yang, R.~Zhang, and J.~Y.~Goulermas, ``Towards better
long-tailed oracle character recognition with adversarial data augmentation,''
\emph{Pattern Recognit.}, vol.~140, Art.~109534, 2023, doi:~10.1016/j.patcog.2023.109534.

\bibitem{yue2022dynamic}
X.~Yue, H.~Li, Y.~Fujikawa, and L.~Meng, ``Dynamic dataset augmentation for deep
learning-based oracle bone inscriptions recognition,'' \emph{ACM J. Comput. Cult.
Herit.}, vol.~15, no.~4, pp.~1--20, 2022.

\bibitem{xia2025mapsam}
X.~Xia, D.~Zhang, W.~Song, W.~Huang, and L.~Hurni, ``MapSAM: Adapting segment anything
model for automated feature detection in historical maps,'' \emph{GIScience \& Remote
Sensing}, vol.~62, no.~1, Art.~2494883, 2025, doi:~10.1080/15481603.2025.2494883.

\bibitem{azzoni2017}
A.~Azzoni, E.~R.~M.~Dusinberre, M.~B.~Garrison, W.~F.~M.~Henkelman, C.~E.~Jones, and
M.~W.~Stolper, ``Persepolis administrative archives,'' \emph{Encyclopaedia Iranica},
online ed., 2017. [Online]. Available:
\url{http://www.iranicaonline.org/articles/persepolis-admin-archive}

\end{thebibliography}

\end{document}